\pdfoutput=1

\documentclass[11pt]{article}

\usepackage[preprint]{acl}

\usepackage{times}
\usepackage{latexsym}
\usepackage{booktabs}
\usepackage{adjustbox}

\usepackage[T1]{fontenc}

\usepackage[utf8]{inputenc}

\usepackage{microtype}

\usepackage{inconsolata}

\usepackage{graphicx}

\usepackage{lipsum}
\usepackage{subcaption}

\title{TakeLab Retriever: AI-Driven Search Engine \\ for Articles from Croatian News Outlets}

\author{David Dukić$^{\dagger}$ \and Marin Petričević \and Sven Ćurković \and Jan Šnajder  \\ TakeLab, Faculty of Electrical Engineering and Computing, University of Zagreb}

\begin{document}
\maketitle
\def\thefootnote{$\dagger$}\footnotetext{Corresponding author: \texttt{david.dukic@fer.hr}}
\renewcommand{\thefootnote}{\arabic{footnote}}

\begin{abstract}
TakeLab Retriever is an AI-driven search engine designed to discover, collect, and semantically analyze news articles from Croatian news outlets. It offers a unique perspective on the history and current landscape of Croatian online news media, making it an essential tool for researchers seeking to uncover trends, patterns, and correlations that general-purpose search engines cannot provide.
TakeLab retriever utilizes cutting-edge natural language processing (NLP) methods, enabling users to sift through articles using named entities, phrases, and topics through the web application.\footnote{\footnotesize Takelab Retriever web application is up and running. You can try it out here \url{https://retriever.takelab.fer.hr}.} This technical report is divided into two parts: the first explains how TakeLab Retriever is utilized, while the second provides a detailed account of its design. In the second part, we also address the software engineering challenges involved and propose solutions for developing a microservice-based semantic search engine capable of handling over ten million news articles published over the past two decades.
\end{abstract}

\section{Introduction}
\label{sec:introduction}

The large volume of news content generated daily and distributed through online outlets far surpasses human cognitive capacity, making it challenging to keep up with all published articles, let alone identify which ones are worth reading.  While general-purpose search engines such as Google, Bing, and DuckDuckGo make accessing news articles easier, they often produce incomplete and biased results. Typically, these engines leave users uncertain about which articles are excluded from the results list, how search results are ranked, and why certain content is prioritized. Furthermore, general-purpose search engines often provide access to only a subset of the articles that are still available on publishers' websites, with no access to the results archive. These limitations are not confined to layman users; they also affect researchers in the social sciences, including political scientists, media analysts, psychologists, and sociologists, who depend on search engine results to analyze news content in their studies. Studies that rely on the results of general-purpose search engines often use samples that are biased, non-random, or too small to be representative. Furthermore, general-purpose search engines often yield different results for the same query. For instance, \citet{lewandowski2015evaluating} showed that Google and Bing produce different search results, particularly when comparing informational and navigational queries, with the difference being more pronounced for navigational queries. Another challenge arises from the focus on developing search engines for widely spoken languages, such as English, often leading to less precise results for lower-resource languages, including Slavic languages like Croatian. In these cases, even when the retrieved article sample is representative, automatically analyzing the content of these articles remains problematic.

To identify the challenges faced by social science researchers who analyze news media content, particularly those engaged in the predominantly manual process of gathering, organizing, and interpreting large volumes of media data, we consulted with researchers who regularly analyze articles from Croatian news outlets, collecting their insights and feedback. Even the tech-savvy researchers relied on results from general-purpose search engines, manually collecting articles and conducting their analyses using spreadsheets. In the best cases, these manually collected articles were analyzed using open-source data mining toolkits such as Orange\footnote{\footnotesize\url{https://orangedatamining.com}} or more specialized but language-limited closed-source tools such as Communalytic.\footnote{\footnotesize\url{https://communalytic.com}} The feedback we received highlighted a clear need for a search engine capable of automatically collecting and semantically analyzing articles from relevant Croatian news outlets. Such a tool would allow researchers to work with comprehensive collections of articles or sample from them, ensuring access to unbiased data without the need for time-consuming manual searches related to their research questions. Equally important, we identified a need for advanced semantic analysis of the articles' content, going beyond basic word frequency methods or general-purpose data mining toolkits.

Motivated by these findings, we developed a specialized semantic search engine, \textbf{TakeLab Retriever}, which we describe in detail in this report. Designed for real-time, automatic retrieval, content extraction, and semantic analysis, TakeLab Retriever is an AI-driven search engine that processes articles from Croatian news outlets using state-of-the-art natural language processing (NLP) models. These include core, lower-level linguistic processing models such as part-of-speech (POS) tagging, tokenization, and dependency parsing models, as well as higher-level models such as the named entity recognition (NER) model, a named entity linking (NEL) model, and a multi-label topic model. The results from these models enhance the quality of semantic search of articles and enable the presentation of articles and by combining various tags assigned to each article, such as named entities, topics, and phrases. 

TakeLab Retriever is unbiased, more precise, and better tailored for Croatian news article retrieval and analysis than general-purpose search engines. As of November 2024, it analyzes a vast archive of ten million news articles, with thousands more added daily. The advanced NLP methods and language models integrated into the search engine enable it to support the precise and unbiased retrieval of information for specific research questions. Moreover, TakeLab Retriever provides an in-depth view of the history and current landscape of online Croatian news, with intuitive visualizations built into a user-friendly web application. By updating and indexing new articles daily, TakeLab Retriever sets a new standard for media research in Croatia, offering deeper insights into trends, patterns, and correlations within online news content. To our knowledge, no similar solutions exist for Croatian news outlets. 




\section{Using the Search Engine}

TakeLab Retriever is primarily intended for researchers (such as political scientists, media analysts, psychologists, and sociologists) who wish to analyze news media content. TakeLab Retriever can be used to analyze comprehensive and unbiased article collections pertinent to specific research questions. The search engine is available to both affiliated and independent researchers for non-commercial use. It has been available to the public since November 2022. Currently (November 2024), the search engine analyzes the content from 33 Croatian news outlets. Table~\ref{tab:portal_counts} shows scraped article counts by news outlets from the database (an additional nine news outlets are crawled but not shown in the web application).

\subsection{Article Collection}

\begin{table}[t!]
\centering
\begin{tabular}{lr}
\toprule
\textbf{News Outlet} & \textbf{Count} \\
\hline
\href{https://www.index.hr}{index.hr} & 1,254,026 \\
\href{https://www.24sata.hr}{24sata.hr} & 1,091,578 \\
\href{https://www.vecernji.hr}{vecernji.hr} & 1,084,002 \\
\href{https://www.jutarnji.hr}{jutarnji.hr} & 986,694 \\
\href{https://www.net.hr}{net.hr} & 964,227 \\
\href{https://www.tportal.hr}{tportal.hr} & 855,193 \\
\href{https://www.dnevnik.hr}{dnevnik.hr} & 703,372 \\
\href{https://www.slobodnadalmacija.hr}{slobodnadalmacija.hr} & 536,334 \\
\href{https://www.glas-slavonije.hr}{glas-slavonije.hr} & 511,595 \\
\href{https://www.narod.hr}{narod.hr} & 369,374 \\
\href{https://www.direktno.hr}{direktno.hr} & 350,978 \\
\href{https://www.rtl.hr}{rtl.hr} & 240,735 \\
\href{https://www.hrt.hr}{hrt.hr} & 231,193 \\
\href{https://www.dnevno.hr}{dnevno.hr} & 211,433 \\
\href{https://n1info.hr/}{hr.n1info.com} & 210,648 \\
\href{https://www.novilist.hr}{novilist.hr} & 187,648 \\
\href{https://www.telegram.hr}{telegram.hr} & 121,872 \\
\href{https://www.h-alter.org}{h-alter.org} & 68,499 \\
\href{https://www.bug.hr}{bug.hr} & 33,806 \\
\href{https://www.priznajem.hr}{priznajem.hr} & 31,607 \\ 
\href{https://www.plusportal.hr}{plusportal.hr} & 31,395 \\
\href{https://www.geopolitika.news}{geopolitika.news} & 26,715 \\
\href{https://www.teleskop.hr}{teleskop.hr} & 21,682 \\
\href{https://www.tris.com.hr}{tris.com.hr} & 15,386 \\
\href{https://www.netokracija.com}{netokracija.com} & 14,303 \\
\href{https://www.lupiga.com}{lupiga.com} & 13,100 \\
\href{https://www.hop.com.hr}{hop.com.hr} & 11,390 \\
\href{https://www.tribun.hr}{tribun.hr} & 8,962 \\ 
\href{https://www.crol.hr}{crol.hr} & 6,231 \\
\href{https://www.paraf.hr}{paraf.hr} & 6,144 \\
\href{https://www.forum.tm}{forum.tm} & 3,981 \\
\href{https://www.liberal.hr}{liberal.hr} & 3,948 \\
\href{https://www.dokumentarac.hr}{dokumentarac.hr} & 477 \\
\bottomrule
\end{tabular}
\caption{Scraped article counts by news outlet (counts for November 2024).}
\label{tab:portal_counts}
\end{table}

\begin{figure*}[t!]
    \centering
    \begin{subfigure}{0.75\linewidth}
        \includegraphics[width=\linewidth]{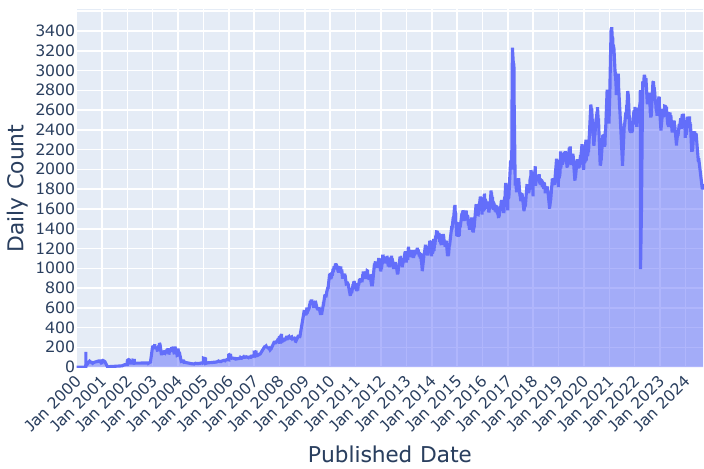}
        \caption{}
        \label{fig:retriever_published_articles}
    \end{subfigure} \\
    \vspace{0.5cm}
    \begin{subfigure}{0.75\linewidth}
        \includegraphics[width=\linewidth]{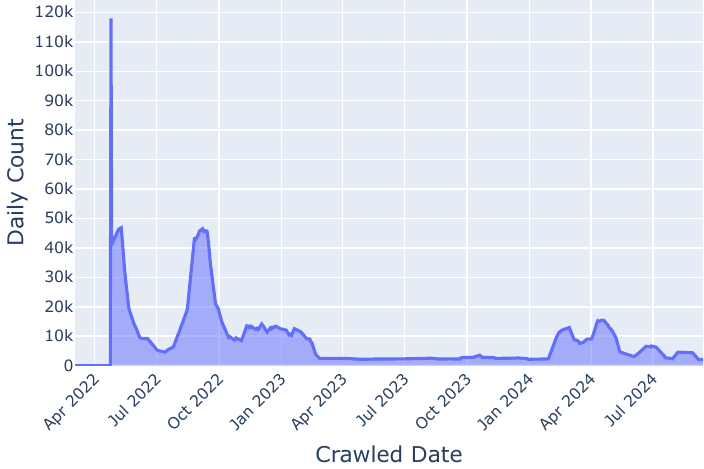}
        \caption{}
        \label{fig:retriever_crawled_articles}
    \end{subfigure}
    \caption{30-day rolling averages of published and crawled articles over time: (a) Rolling average number of published articles from news outlets within 12-month periods (starting with January 2000); (b) Rolling average number of crawled articles from news outlets within 3-month periods (from last two years).}
    \label{fig:retriever_published_crawled_articles}    
\end{figure*}

With TakeLab Retriever, we located and processed more than ten million unique article URLs on 42 regional news outlets. Figure~\ref{fig:retriever_published_articles} shows the 30-day rolling average number of published articles over time. We discovered articles from the early 2000s until today (November 2024). The counts suggest that more content is being published online as the years go by. However, as we move into the history of news outlets, the number of discovered articles drops. While this can be partially explained by regional outlets publishing less content two decades ago, the increasing trend is mostly an artifact of how the articles are collected. Specifically, we collect new articles by following links from those we have already visited. These links become increasingly sparse as we move back in time, as newer articles rarely link to older ones, making it more difficult to locate and retrieve archived content. Starting with articles published from 2018 onwards, we can locate, on average, between 2000 and 3000 articles published each day. 

We started crawling the content intensively in 2022, starting from the seed URLs set with over 100k article URLs we collected using an earlier version of the search engine. Figure~\ref{fig:retriever_crawled_articles} shows our crawling dynamics with 30-day rolling averages starting in 2022. Since October 2022, when the search engine stabilized after processing the initial seed set, we have been crawling and processing between 2,000 and 15,000 articles daily. The fluctuations (variations and spikes) in the daily crawled counts arise from several factors. Downtime in the crawling system disrupts continuous data collection. Similarly, adding support for more outlets alters the system's crawling dynamics. Finally, articles with incorrect dates or incorrectly extracted dates can create artificial spikes in the data.

\subsection{Semantic Search}

The main feature of the TakeLab Retriever web application is its ability to retrieve all articles that match a user-provided query. The query can include semantic constraints (i.e., referring to the semantic content of articles, such as named entities or topics mentioned) and can be as simple or as complex as needed. More complex queries can be constructed using Boolean algebra operators by combining one or more of the following constraints:

\begin{itemize}
    \item Which news outlet is the article on (user can select all news outlets or only specific ones)?
    \item Which named entity/entities are (or are not) mentioned in the article?
    \item Which arbitrary phrase(s) are (or are not) mentioned in the article?
    \item Which topic(s) does (or does not) the article belong to?
    \item Is the article automatically classified as a low-quality article (an article that is likely not a proper news story, including very short articles, opinion pieces, astrology content, or photo galleries)?
\end{itemize}

This functionality makes it seamless to locate and analyze text data published on Croatian news outlets while delivering high-quality analyses. Users can explore trends with just a few clicks and keystrokes through the web application without needing advanced technical expertise.

\begin{figure*}
    \centering
    \includegraphics[width=0.8\linewidth]{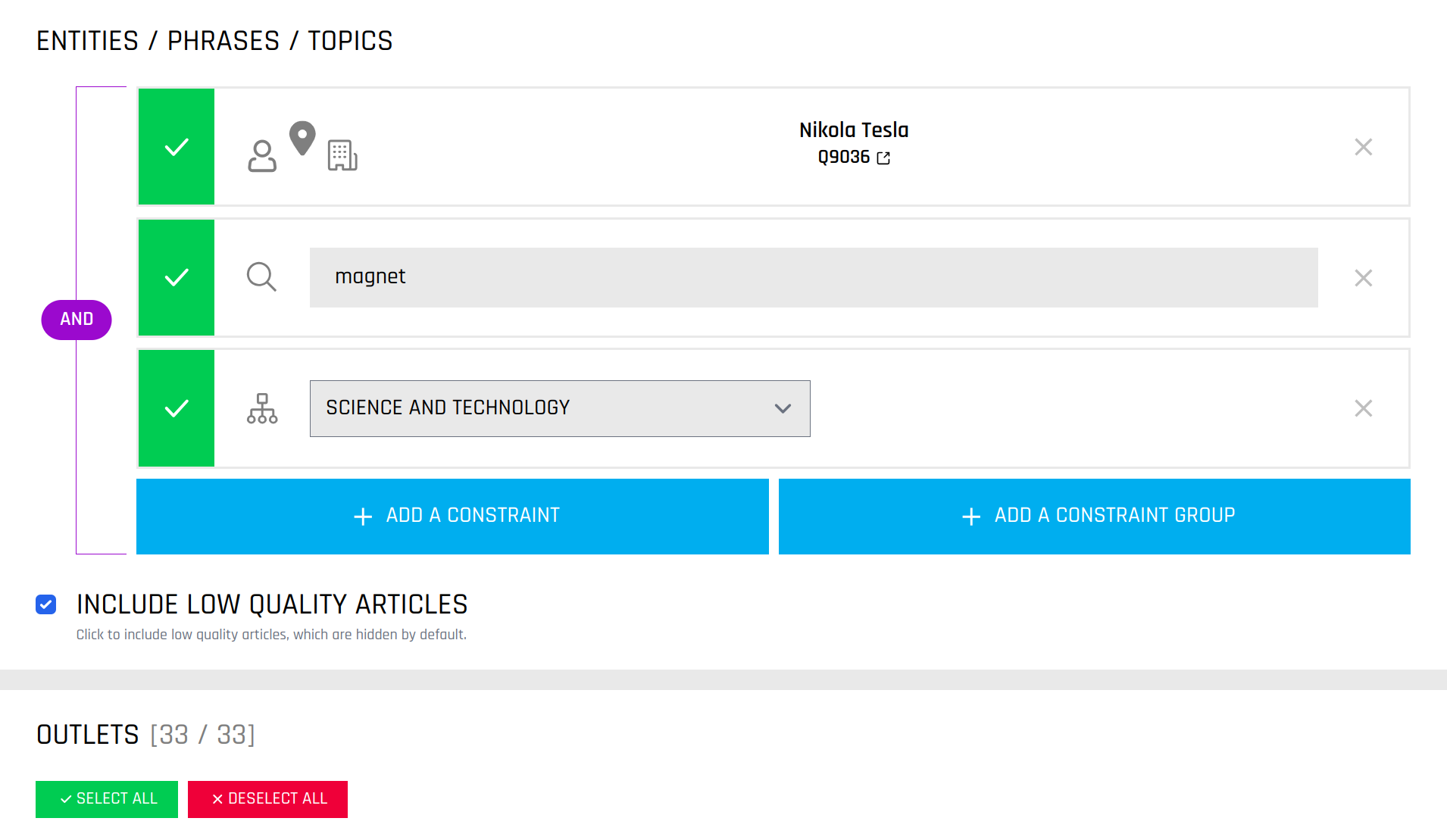}
    \caption{An example of a complex query in TakeLab Retriever web application with combinations of different search constraints.}
    \label{fig:search_example}
\end{figure*}

\begin{figure*}
    \centering
    \includegraphics[width=1.0\linewidth]{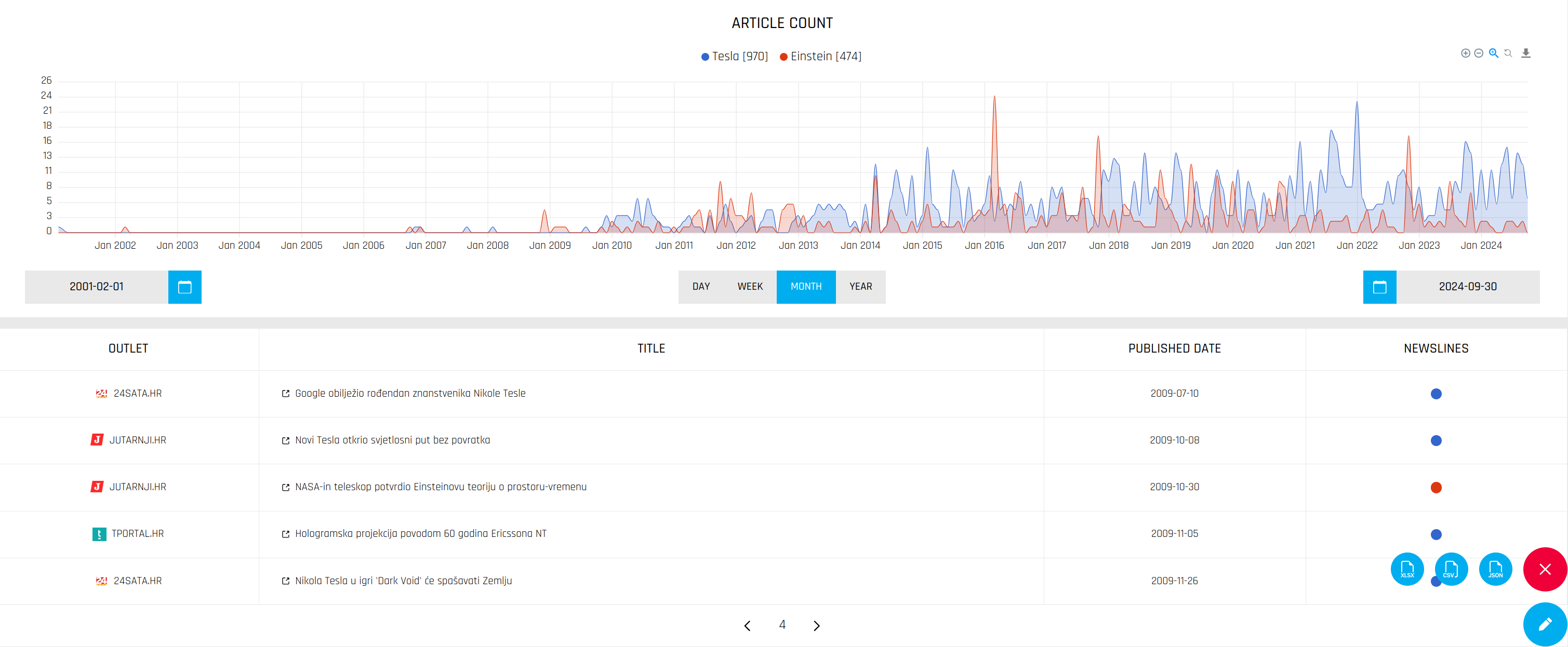}
    \caption{TakeLab Retriever web application search result for \emph{Nikola Tesla} and \emph{Albert Einstein} entities constraint in combination with the topic \emph{SCIENCE AND TECHNOLOGY} on 33 Croatian news outlets with no articles hidden from the results.}
    \label{fig:retriever_search_result}
\end{figure*}

Figure~\ref{fig:search_example} shows an example of a complex query that runs on 33 news outlets, does not omit low-quality articles from the result list, and uses the Boolean operator AND between the following search constraints: \emph{Nikola Tesla} as an entity constraint, \emph{magnet} as a phrase constraint, \emph{SCIENCE AND TECHNOLOGY} as a topic constraint. The result is returned as a summary of article metadata and statistics, shown to the user as a graph with the number of articles in time and the table with metadata (cf.~Figure~\ref{fig:retriever_search_result}). The user can zoom in and out on the article graph, and these actions update the metadata table. Users can also export the article metadata in either XLSX, CSV, or JSON format for a closer look into the article URLs and metadata. 

Figure~\ref{fig:retriever_search_result} shows two separate search results (two sets of constraints) on the same graph. Each constraint over the articles invokes a line on the graph, which we call a \emph{Newsline}. A set of constraints with entity \emph{Nikola Tesla} and topic \emph{SCIENCE AND TECHNOLOGY} invokes the first Newsline, which we called \emph{Tesla}. Similarly, the set of constraints with entity \emph{Albert Einstein} and topic \emph{SCIENCE AND TECHNOLOGY} invokes the second Newsline, which we named \emph{Einstein}. These constraint combinations are arbitrary and could have been more or less strict. Combining constraints arbitrarily and defining multiple isolated constraints enables users to unveil trends, patterns, and correlations in time quickly and easily. For example, based on the absolute number of article mentions (\emph{article count}), Nikola Tesla seems more popular than Albert Einstein in Croatian news outlets.

\section{Search Engine Design}
\label{sec:search_engine_design}

\begin{figure*}
    \centering
    \includegraphics[width=1.0\linewidth]{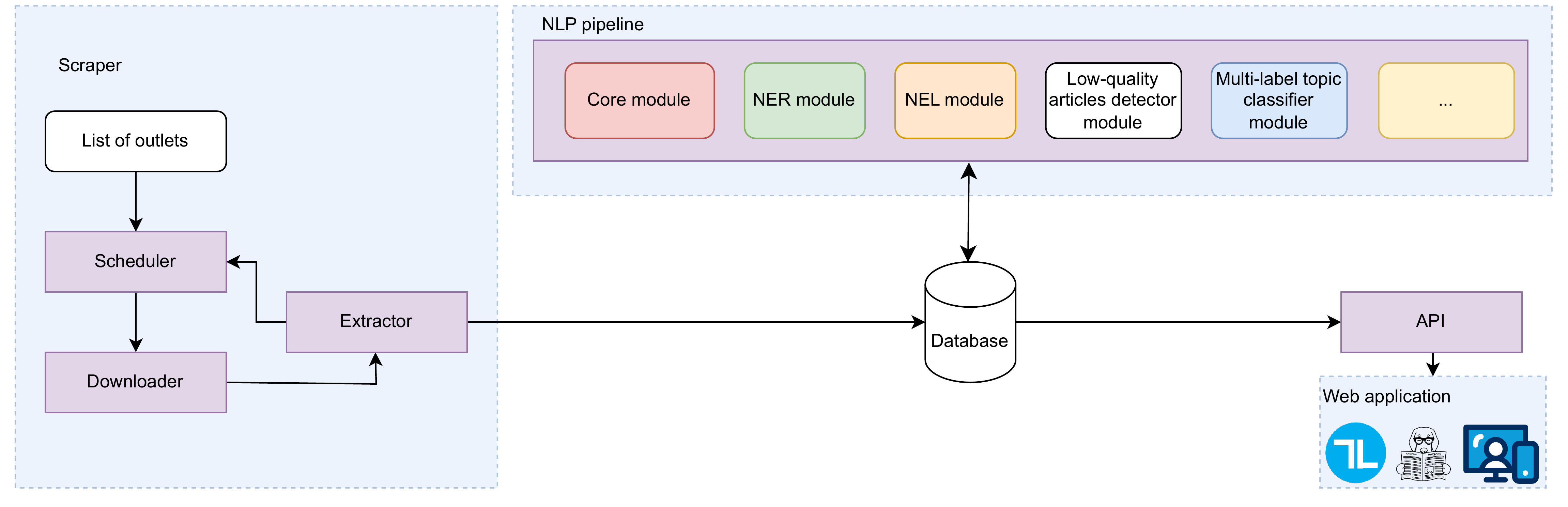}
    \caption{The architecture of TakeLab Retriever.}
    \label{fig:retriever_arch}
\end{figure*}

The TakeLab Retriever search engine comprises three parts: the scraper, the NLP modules, and the web application, where the application and the database communicate through the API. Figure~\ref{fig:retriever_arch} shows the architecture of the search engine. The design was guided by a number of key requirements:

\begin{itemize}
    \item{asynchronous I/O},
    \item{concurrent processing pipelines},
    \item{flexible data flow},
    \item{independent and trivial scaling of the parts of the search engine},
    \item{high robustness},
    \item{high performance}.
\end{itemize}

We adopted a microservice architecture to address these requirements, where each service is run separately in its container. Containers are defined, created, started, and managed with Docker.\footnote{\footnotesize\url{https://www.docker.com}} 
We opted for PostgreSQL\footnote{\footnotesize\url{https://www.postgresql.org}} as a message broker for service communication. Specifically, we implemented non-blocking priority queues using PostgreSQL tables that allow arbitrary JSON messages. This enables us to prioritize particular articles as they move through the search engine (for example, after updating a specific NLP model, we want to run inference on all articles in the database but prioritize new ones). PostgreSQL could handle both the read/write-intensive queue operations and the read-heavy tasks associated with managing collected data. Its ACID (atomicity, consistency, isolation, and durability) properties further enhance the search engine's robustness by ensuring resilience to data loss. To increase reliability, we implemented an error queue for each component, where failed tasks are sent for later review by a human or automatically retried after a predefined delay. A notable advantage of this microservice and queue-oriented architecture is the ease of rerunning tasks. Simply pushing the relevant data from the database back into the appropriate queue allows us to effortlessly reprocess tasks as needed. To keep track of the search engine state, we implemented monitoring, logging, and alerting solutions using a combination of InfluxDB,\footnote{\footnotesize\url{https://github.com/influxdata/influxdb}} Telegraf,\footnote{\footnotesize\url{https://github.com/influxdata/telegraf}} and Grafana.\footnote{\url{https://github.com/grafana/grafana}}

\subsection{The Scraper}

The scraper is responsible for locating the articles, downloading them, and storing the scraped article content in the database. We opted for a custom solution without relying on any scraping framework, adopting outlet-agnostic data collection and content extraction. We identified common patterns across different news outlets and standardized the crawling process.


Our scraper operates as a closed-loop system (cf.~Figure~\ref{fig:retriever_arch}) composed of three main components: the scheduler, the downloader, and the extractor. Each news outlet is assigned its crawler, which starts with a list of initial URLs (i.e., the outlet's homepages) and a set of regular expressions (regexes). These regex patterns determine which URLs should lead to articles, which should be excluded, and which should be entirely ignored to prevent unnecessary expansion down the URL structure. During our initial news outlet analysis, we discovered that each news outlet's article webpages follow distinct URL patterns that can be effectively identified using a regex. We also unveiled that most news outlets follow a similar structure: homepages with links to articles, where the article URLs follow predictable patterns. To maximize user engagement, news outlets interlink articles, creating a hidden web of article connections within each news outlet page. By leveraging this structure, we built a search engine that periodically visits news outlet homepages, crawls article URLs, and recursively discovers additional articles, ensuring comprehensive content coverage.

\subsubsection{The Scheduler}

The scheduler component serves as the entry point to our search engine. It receives URLs extracted by the extractor, checks whether they have already been visited, and, if not, assigns them to their respective downloader queues (each crawler has its own queue). If the URL has not yet been visited, it is added to the visited URLs set, which is stored in PostgreSQL as a single-column table.\footnote{\footnotesize{As of November 2024, we have over 80 million visited URLs from the news outlets we continuously scrape.}}  
This allows us to distinguish between articles within each news outlet based on their URL structure.

The scheduler also includes a mechanism for deferring sending URLs to the downloader queues. This is implemented via a PostgreSQL table with a column that specifies when the URL will be sent to its respective downloader queue. This mechanism is primarily used for recrawling news outlets' homepages at fixed intervals. However, it could be extended to recrawl articles at determined intervals, which would be helpful for live articles that are frequently updated, such as live election coverage. Despite this potential, we have opted not to scrape live articles periodically due to the rarity of such content in regional news outlets. Furthermore, the overhead required to periodically crawl all articles to detect updates in a small fraction is resource-intensive. It is also impossible to differentiate between regular, live, or test articles that may be published unofficially somewhere inside the news outlet link structure but not linked on the news outlet’s homepage.

Additionally, the scheduler assigns a priority to each URL, determining its place in the processing pipeline. This priority propagates throughout the search engine, allowing focus on more relevant and recent content. For example, we prioritize homepages and the articles directly linked from them over archival articles buried deeper within the news outlet's link structure. This ensures near real-time data collection for newly published articles.

\subsubsection{The Downloader}

The downloader component retrieves URLs from the downloader queues, downloads the HTML content, and passes it to the extractor. It adheres to each site's \emph{robots.txt} file, and if none is specified, it ensures a respectful delay between consecutive download requests for the same news outlet. Robust error handling and retry mechanisms are in place to address the many edge cases that can arise with network requests. Given the nature of scraping, the downloader spends significant time waiting for and between network requests. To optimize this, we run each crawler in its green thread (i.e., multiple threads scheduled by a user-space runtime running in a single operating thread), utilizing Python’s native AsyncIO library.  
Recently, some news outlets started locking article content under paywalls. We tackled this by purchasing the yearly subscription for each such news outlet and opening an AIOHTTP\footnote{\footnotesize{\url{https://docs.aiohttp.org/en/stable}}} client session for scraping the content of that news outlet through the downloader component.

\subsubsection{The Extractor}

The extractor component processes the HTML received from the downloader, extracting relevant information and populating the columns in the PostgreSQL \emph{article} table. This table is the main table in the database for storing all article data and metadata, such as the article body, title, and published date, which are essential for subsequent processing, data visualization, and presentation in the web application. This table also hosts the results of applying NLP models to the extracted title and body. The extractor ensures the data is accurate and clean, aiming to reduce redundancy through deduplication against previously scraped articles. Once the extraction is complete, the data is saved to the database, and a request is sent to the NLP pipeline queue for further analysis. A critical requirement for the extractor is to clean and normalize the data, ensuring that no other search engine component has to manage this task. For extracting article content such as the body, title, and published date from raw HTML, we use the Trafilatura library \citep{barbaresi-2021-trafilatura}, enforcing the precision mode. Our experiments with extracted data from the articles showed that the recall mode of Trafilatura extraction often pulls in excessive surrounding content, such as advertisements or user comments. To ensure consistency, the extracted title and body are normalized to UTF-8's NFKC form. Except for high-quality article data and metadata extraction, the extractor's responsibility is spreading down the news outlet's link structure since it extracts all the URLs from the HTML, forwarding them to the scheduler (cf.~Figure~\ref{fig:retriever_arch}).  

During the experimentation phase, we observed that the same articles were being recrawled multiple times, even though we tracked visited URLs through the scheduler component. This issue stemmed from three causes: (1) different URLs redirect to the same article (e.g., a URL with an ID form and a slug form), (2) the URL for an article changes due to its title changing, which in turn changes its slug, and (3) URLs having unnecessary (i.e., not required for article identification) query parameters (e.g., tracking information) and fragments. To address this recrawling issue, we implemented a deduplication algorithm to determine if an article already exists in the database. The process begins with a URL-based check to see if an article with the same URL is stored. If this is the case, the existing entry is updated (unless the newly extracted body is too short compared to the old body).\footnote{\footnotesize{We replace the extraction for the already saved URL only if the newly extracted body is bigger than 50\% of the old saved article body length to battle data loss during re-download or re-extraction process.}} If the URL is novel, we perform content-based checks. Specifically, we compute the SimHash \cite{sadowski2007simhash} of the article. This SimHash is based on trigrams of the concatenated alphanumeric characters from both the title and body of the article. We then search the database for exact matches to this SimHash (i.e., with a Hamming distance of 0) and save the article as a new entry if there is no exact SimHash match between the saved articles from the news outlet and the article to be saved.\footnote{\footnotesize{We do not compare SimHashes between news outlets.}} However, if a match exists, the entry in the database is updated, and we keep the shorter URL. This deduplication approach using exact SimHash match serves two purposes: (1) since articles are generally lengthy, SimHash performs well since the impact of the small changes is minimal in comparison with the overall text size and does not alter the SimHash, and (2) exact matching is extremely fast in PostgreSQL when the SimHash column is indexed, whereas searching for approximate matches would require more complex structures such as BK-trees, which would increase TakeLab Retriever's complexity and negatively affect performance. We evaluated this approach (with a Hamming distance of 0) on an automatically collected dataset of approximately 150,000 distinct\footnote{URL was used to determine uniqueness, which is a good approximation when scraping over a short period.} articles and found only two false positives. 

\subsection{The NLP Pipeline}

Once the article passes through the scraper component and is saved to the database, it is sent to the NLP pipeline component for further processing. The NLP pipeline component processes articles sequentially using individual NLP modules. The NLP pipeline component handles the entire pipeline sequentially (i.e., on a single thread). To speed up processing, we can run multiple pipeline containers on available GPU units, which is particularly useful for tasks like reprocessing large parts of the database with updated NLP models.\footnote{\footnotesize{On a day-to-day basis, we run one NLP pipeline instance on each physical GPU unit.}} The NLP pipeline is the only component of TakeLab Retriever that utilizes GPUs. 

\begin{figure*}[!htb]
    \centering
    \includegraphics[width=\linewidth]{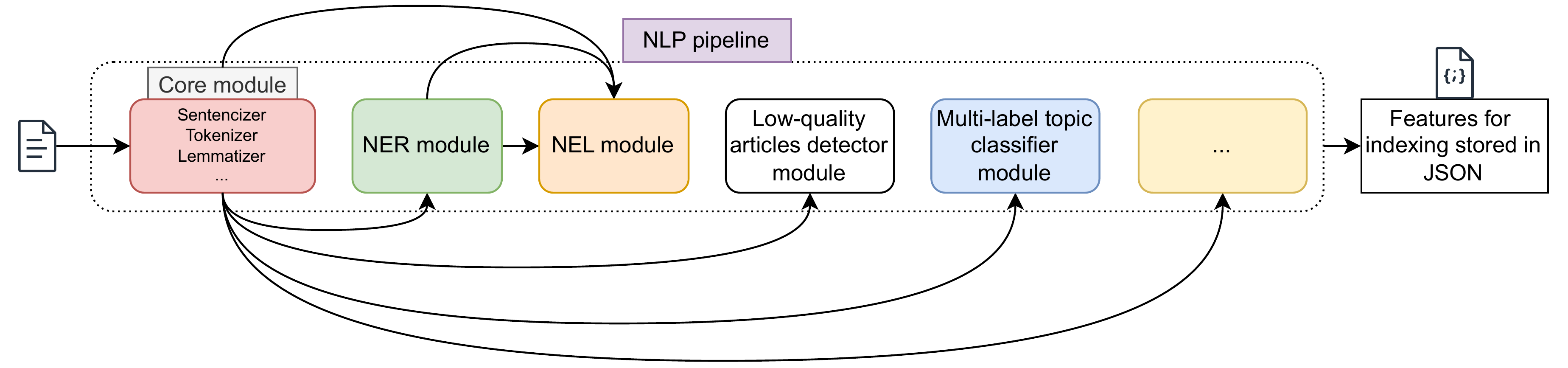}
    \caption{The directed acyclic graph of the NLP pipeline in TakeLab Retriever used to semantically index articles with phrases, entities, and topics. The arcs depict which module depends on the previous ones in the pipeline.}
    \label{fig:retriever_nlp_pipeline}
\end{figure*}

The NLP processing pipeline can be represented as a directed acyclic graph where nodes represent NLP modules (i.e., processors that fetch saved article data and apply NLP model(s)) and edges represent dependencies (i.e., a module may rely on others to complete before it can start), as shown in Figure~\ref{fig:retriever_nlp_pipeline}. Each NLP module in a pipeline defines a processing logic over article entries (whether the models operate on title, body, or both and how) using one or more NLP models. This pipeline design allows for flexible execution: we can update and rerun specific modules (e.g., when a model inside a module is updated) and only rerun those modules that directly or indirectly depend on the updated module, ensuring that all calculated features remain up to date while minimizing computational costs. Having a reindexing support is very useful. When a bug needs to be fixed, or new components are added, there is no need to reprocess the whole archive with all modules in the NLP pipeline. Processing can instead begin from a specific point in the pipeline and continue to the end. We chose a sequential execution model over a single task-per-module approach because the pipeline is not highly parallelizable. The added complexity, particularly the loss of atomicity and increased communication overhead between tasks would outweigh the benefits and potentially degrade performance. Each module processes the article data and returns one or more features, which are stored in a schemaless JSON format in the database in the \emph{features} column of the \emph{article} table. This JSON contains all features calculated for a specific article, with each feature comprising two fields: the data itself and a version number. The version number increments whenever the model for that feature is updated, enabling us to track which articles have been processed with the latest models and which have not. 

Our current NLP pipeline hosts the following modules:
\begin{enumerate}
    \item The \emph{core} module performs core, lower-level, linguistic processing tasks (e.g., sentence segmentation, tokenization, and lemmatization);
    \item The \emph{NER} module extracts named entities from the article;
    \item The \emph{NEL} module links entities with their unique IDs in the knowledge database;
    \item The \emph{low-quality articles detector} module decides if the scraped article is worth reading and if it should be hidden by default based on the quality of its content;
    \item The \emph{multi-label topic classifier} module assigns one or more topics to each article based on the revised IPTC taxonomy.\footnote{\footnotesize\url{https://iptc.org}}
\end{enumerate}

\subsubsection{The Core Module}

The core module is the prerequisite for any other NLP module in the pipeline. This module performs core NLP tasks using the Croatian\footnote{\footnotesize\url{https://spacy.io/models/hr}} spaCy model \verb|hr_core_news_lg| \citep{spacy}. We use the statistical spaCy sentencizer model to split articles into sentences.\footnote{\footnotesize{Our analyses showed that it works better than rule-based sentence segmentation approaches.}} After performing sentence segmentation, we apply the spaCy model for tokenization, POS tagging, morphological features extraction, and dependency parsing. Furthermore, we apply lemmatization with the MOLEX dictionary-based lemmatizer \citep{vsnajder2008automatic} because, unlike the spaCy lemmatizer, it is stable and produces the same lemmas for the same input text, which is better suited for search. We use MOLEX lemmas for our phrase search feature in the web application (every phrase entered for search is lemmatized and matched with lemmatized entries in the database). Finally, through the core module, we instantiate a fastText\footnote{\footnotesize\url{https://fasttext.cc}} Croatian model that other NLP modules in the pipeline depend on. The results of the core module are saved in the database as raw features.

\subsubsection{The NER Module}

The NER module is based on the BERTić encoder model \citep{ljubesic-lauc-2021-bertic} fine-tuned for NER.\footnote{\footnotesize\url{https://huggingface.co/classla/bcms-bertic-ner}} We extract named entities from the article title and body and save each entity's start position, end position, and NER type (location, person, organization, miscellaneous) as raw features. Since we tokenize the text with spaCy and apply the NER model, which depends on BERTić tokenizer, we align the tokens with the Python library tokenizations,\footnote{\footnotesize\url{https://github.com/explosion/tokenizations}} and save NER extractions so that they correspond to spaCy tokens. Therefore, the NER module depends on the core module (cf.~Figure~\ref{fig:retriever_nlp_pipeline}). NER extractions are not exposed to the end user. They are a prerequisite for the NEL module, which is exposed to the user via the web application.

\subsubsection{The NEL Module}

The NEL module is a custom model that matches extracted named entity spans in text with their unique IDs from a knowledge database of entities. We opted for the Wikidata knowledge database due to its rich and continuously updated entity collection.\footnote{\footnotesize{\url{https://www.wikidata.org/wiki}}} NEL depends on the successful execution of both the core module and the NER module for an article.

The NEL module creation involved several steps. First, we trained graph embeddings using PyTorch BigGraph \cite{lerer2019pytorch} on the complete Wikidata graph. We then computed PageRank scores \citep{page1999pagerank} for all items and classified them into three categories—organizations, locations, and persons—using ``instance of'' and ``subclass of'' relations as defined in \cite{shanaz2019named}. We filtered these items to include only those with Croatian labels (including BCMS language variants, excluding Cyrillic script). The filtered items, their embeddings, PageRank scores, and all Wikidata labels and aliases are stored in our database. 

For entity linking, the system first retrieves candidate entities for each NER-extracted mention by matching lemmatized mentions against stored labels and aliases. The initial solution assigns each mention to the candidate with the highest PageRank score. This solution is then iteratively refined (maximum five iterations, though convergence typically occurs within two to three steps) by computing similarity scores between candidates and current best solutions for all other mentions. For each mention, we calculate the dot product between its candidates' embeddings and the embeddings of the current best solution for other mentions, selecting the candidate with the highest aggregate similarity score as the new solution.


This approach works well for well-known entities but sometimes fails for less-known entities. Nevertheless, it enables users to search for entities through the web application and to verify using Wikidata ID that the entity they are searching for is the right one.\footnote{\footnotesize{For example, if the user searches for named entity \emph{Tesla}, the user can be sure based on the ID 9036 (\url{https://www.wikidata.org/wiki/Q9036}) that the search is meant to retrieve the article mentioning the famous inventor and not a company, which has a different ID.}} The alternative approach we plan to integrate into the system is a matching model, which was trained to embed named entities and predict their ID in the knowledge database directly.

\subsubsection{The Low-Quality Articles Detector Module}
We noticed that a significant portion of the articles we collected with the scraper contained a low number of tokens or were not pieces of text a reader would consider a genuine article (opinion pieces, astrology content, photo galleries, etc.). However, these low-quality articles could not be spotted algorithmically since their URL schema was identical to high-quality articles. Therefore, we decided to build a statistical classifier to predict if the articles are of such low quality that we should hide them from the search results to enhance user experience on our semantic search engine. Our team annotated a sample of a couple hundred articles by declaring articles low-quality if the article was \emph{not worth reading}. We trained a support vector machine (SVM) binary classifier on the annotated data that classifies the articles relying on the fastText embedding \citep{bojanowski-etal-2017-enriching} of their concatenated title and body. We leverage the Croatian fastText embedding model for this purpose. Articles are ultimately assigned \emph{low-quality} or \emph{not-low quality} labels. This module also relies on successfully executing the core module in the pipeline. If the article (title + body) is shorter than 50 tokens, it is automatically considered of low-quality and hidden from the search results. If the number of tokens exceeds 50, it is hidden only if the low-quality articles detector module labels it as low-quality. In the web application, users can either include or remove low-quality articles from the presentation of the results. Our database hosts 11.83\% articles hidden by default in the web application, making up a significant portion of the total number of crawled news articles.


\subsubsection{The Multi-Label Topic Classifier Module}

To assign topics to articles, we developed a multi-label topic classifier module, which assigns one or more predefined topics to the article based on the title and body of the article. For this purpose, our team annotated a sample of almost a thousand articles with 17 topics based on the revised IPTC taxonomy. Table~\ref{tab:iptc_topics} provides a breakdown of the articles in our database by their assigned topics.\footnote{\footnotesize{Note that each article can have several topics assigned.}}

\begin{table}[!htb]
    \centering
    \begin{adjustbox}{width=\columnwidth}
    \begin{tabular}{lr}
    \toprule
        \textbf{IPTC topic} & \textbf{Count} \\ \hline
        SPORT & 1,972,713 \\ 
        HOBBY AND PERSONAL INTEREST & 1,964,704 \\ 
        POLITICS & 1,367,163 \\ 
        CRIME, LAW AND JUSTICE & 971,341 \\ 
        ECONOMY, BUSINESS AND FINANCE & 911,022 \\ 
        DISASTER, ACCIDENT AND EMERGENCY INCIDENT & 669,387 \\ 
        HEALTH & 634,151 \\ 
        ARTS, CULTURE, ENTERTAINMENT AND MEDIA & 467,566 \\ 
        CONFLICTS, WAR AND PEACE & 360,649 \\ 
        SCIENCE AND TECHNOLOGY & 332,232 \\ 
        ENVIRONMENT & 217,948 \\ 
        EDUCATION & 153,699 \\ 
        WEATHER & 148,494 \\ 
        RELIGION & 116,417 \\ 
        LIFESTYLE AND LEISURE & 96,762 \\ 
        LABOUR & 88,905 \\ 
        SOCIETY & 14,053 \\ 
    \bottomrule
    \end{tabular}
    \end{adjustbox}
    \caption{List of revised IPTC topics ordered by counts of over ten million scraped news articles (counts for November 2024).}
    \label{tab:iptc_topics}
\end{table}

On top of annotated data, we trained the Omikuji classifier,\footnote{\footnotesize\url{https://github.com/tomtung/omikuji}} an efficient implementation of partitioned label trees \citep{prabhu2018parabel}, which assigns one or more topics to each article and works well even for rare multi-label combinations in the training data. Each article is embedded with the fastText model for Croatian to enable high-quality classification. The multi-label topic classifier module also depends on the core module. This module allows users to search for articles based on the assigned predefined topics via the web application. 

\subsection{API and Web Application}

We do not provide direct access to our data or stored features. Instead, we offer specific querying endpoints through a RESTful API tailored to our target audience. Since we do not expect our users to have technical expertise, they will primarily interact with the semantic search engine via a web application frontend, which handles all server communication and data display. This approach improves the user experience. Each frontend query is translated through backend logic into an SQL query, executed over the database by sending a request to a specific API endpoint. Issuing a query like the one shown in Figure~\ref{fig:retriever_search_result} is fast because each search constraint (entities, phrases, or topics) has a dedicated column in the \emph{article} table of the database, with either a B-tree or GIN index applied to optimize query performance. We also perform caching for a faster user experience and use materialized view, speeding up entity count over articles. The web application is written in Vue.js\footnote{\footnotesize{\url{https://vuejs.org}}} and Tailwind CSS.\footnote{\footnotesize{\url{https://tailwindcss.com/}}} While we plan to release API endpoints for more advanced users, this feature is scheduled for one of the future updates of TakeLab Retriever.

\section{Conclusion}
\label{sec:conclusion}

TakeLab Retriever search engine offers a comprehensive and tailored solution for automated scraping, retrieval, semantic analysis, and presentation of news articles from Croatian online news outlets. By integrating state-of-the-art NLP models and adopting a robust microservice architecture, TakeLab Retriever overcomes the limitations of general-purpose search engines and streamlines the research process for scholars in the social sciences and beyond. The ability of TakeLab Retriever to automatically unveil and semantically index over ten million articles, combined with its web application, enables fast article retrieval and improves the quality and precision of research. With its ongoing updates and daily indexing of new articles, TakeLab Retriever sets a new standard for media research in Croatia, contributing to a more profound understanding of trends, patterns, and correlations in online news content. 

For future work, we intend to improve the web application experience and introduce a concept of the user where user-specific queries could be saved into a user account, trending queries could be detected automatically, and queries could be shared between users to ensure result reproducibility. Furthermore, we plan to extend the semantic search engine with the new NLP modules, including the keyphrase extraction module, the semantic similarity module, and a module for sentiment analysis in article titles. Finally, we wish to extend the scraper to cover more Croatian news outlets as well as regional news outlets. 

\section*{Acknowledgments}

We extend our heartfelt thanks to everyone involved in creating and developing the TakeLab Retriever search engine. We are particularly grateful to Ivan Krišto, whose software engineering expertise and mentorship were instrumental in achieving search engine stability. The engine has evolved to its current state through the valuable contributions of numerous interns, master's and doctoral students, and social science researchers.




\bibliography{anthology,custom}

\end{document}